\documentclass[journal,twoside,web]{ieeecolor}
\usepackage{generic}
\usepackage{cite}
\usepackage{amsmath,amssymb,amsfonts}
\usepackage{algorithmic}
\usepackage{graphicx}
\usepackage{algorithm,algorithmic}
\usepackage{hyperref}
\hypersetup{hidelinks=true}
\usepackage{textcomp}

\usepackage{caption} 
\usepackage{float}
\usepackage{booktabs}
\usepackage{makecell}
\usepackage{multirow}
\def\BibTeX{{\rm B\kern-.05em{\sc i\kern-.025em b}\kern-.08em
    T\kern-.1667em\lower.7ex\hbox{E}\kern-.125emX}}
\begin{document}
\title{DistilCLIP-EEG: Enhancing Epileptic Seizure Detection Through Multi-modal Learning and Knowledge Distillation}
\author{Zexin Wang, Lin Shi, Haoyu Wu, Junru Luo, Xiangzeng Kong and Jun Qi, \textit{Member, IEEE}
\thanks{Manuscript received 50 August 2024; revised 50 January 2024 and 50 April 2024; accepted 50 May 2024. Date of publication 50 May 2024; date of current version 50 August 2024. This work was supported in part by National Natural Science Foundation of China (62301452), in part by the Xi'an JiaoTong-Liverpool University Research Development Fund (RDF-21-01-080), in part by Jiangsu Province Digital Twin Technology Engineering Research Center for Key Equipment of Petrochemical Process (Su Fa Gai Gao Ji Fa [2019] No.1125), and in part by Changzhou Key Laboratory of Urban Big Data Analysis and Application Technology (CM20193007). (Corresponding author: Jun Qi.)}
\thanks{Zexin Wang and Lin Shi are with the Aliyun School of Big Data, Changzhou University, China. The authors contribute equally}
\thanks{Junru Luo is with the Aliyun School of Big Data, Changzhou University, China.}
\thanks{Haoyu Wu is with the Department of Computing, Xi'an JiaoTong-Liverpool University, China, and also with the Department of Computer Science, University of Liverpool, UK.}
\thanks{Xiangzeng Kong is with the Center for Artificial Intelligence in Agriculture, Fujian Agriculture and Forestry University, China.}
\thanks{Jun Qi is with the Department of Computing, Xi'an JiaoTong-Liverpool University, China (e-mail: jun.qi@xjtlu.edu.cn).}}

\maketitle
\begin{abstract}
Epilepsy is a prevalent neurological disorder marked by sudden, brief episodes of excessive neuronal activity caused by abnormal electrical discharges, which may lead to some mental disorders. Most existing deep learning methods for epilepsy detection rely solely on unimodal EEG signals, neglecting the potential benefits of multimodal information. To address this, we propose a novel multimodal model, DistilCLIP-EEG, based on the CLIP framework, which integrates both EEG signals and text descriptions to capture comprehensive features of epileptic seizures. The model involves an EEG encoder based on the Conformer architecture as a text encoder, the proposed Learnable BERT (BERT-LP) as prompt learning within the encoders. Both operate in a shared latent space for effective cross-modal representation learning. To enhance efficiency and adaptability, we introduce a knowledge distillation method where the trained DistilCLIP-EEG serves as a teacher to guide a more compact student model to reduce training complexity and time. On the TUSZ, AUBMC, and CHB-MIT datasets, both the teacher and student models achieved accuracy rates exceeding 97\%. Across all datasets, the F1-scores were consistently above 0.94, demonstrating the robustness and reliability of the proposed framework. Moreover, the student model's parameter count and model size are approximately 58.1\% of those of the teacher model, significantly reducing model complexity and storage requirements while maintaining high performance. These results highlight the potential of our proposed model for EEG-based epilepsy detection and establish a solid foundation for deploying lightweight models in resource-constrained settings.
\end{abstract}


\section{Introduction}
\label{sec:introduction}
\IEEEPARstart{E}{pilepsy} is a chronic neurological disorder that affects millions of people worldwide, and accurate seizure detection is crucial for timely medical intervention \cite{WHO}. In addition to the physiological impact of epilepsy, the disorder affects the mental health of patients. Patients with epilepsy are at significantly higher risk of facing mental health challenges such as anxiety, depression, and reduced quality of life. The unpredictability of seizures often leads to increased stress and social isolation. 

The traditional diagnosis of epilepsy is to use brain imaging techniques such as magnetic resonance imaging (MRI) \cite{MRI}, functional MRI (fMRI) \cite{fMRI}, and positron emission tomography (PET) \cite{PET}, as these modalities offer high spatial resolution and detailed anatomical information, making them effective for identifying structural abnormalities in the brain. However, they come with several limitations, particularly in real-time monitoring and dynamic analysis of brain activity. MRI and fMRI, for example, suffer from low temporal resolution, making it difficult to capture the fast neuronal activity associated with epileptic seizures. In addition, these imaging techniques are often expensive, require specialized equipment, and are not portable, limiting their accessibility for continuous monitoring in clinical or daily settings. 

This restricts their use in continuous monitoring and in day-to-day contexts, where psychological and social factors play a crucial role. Consequently, there is a growing interest in developing accessible and non-invasive methods for seizure detection that not only provide accurate diagnostic information but also address the psychological well-being of patients, helping to reduce the burden of epilepsy on mental health and improve quality of life.

\begin{figure*}[!t] 
    \centering
    \includegraphics[width = 1\linewidth]{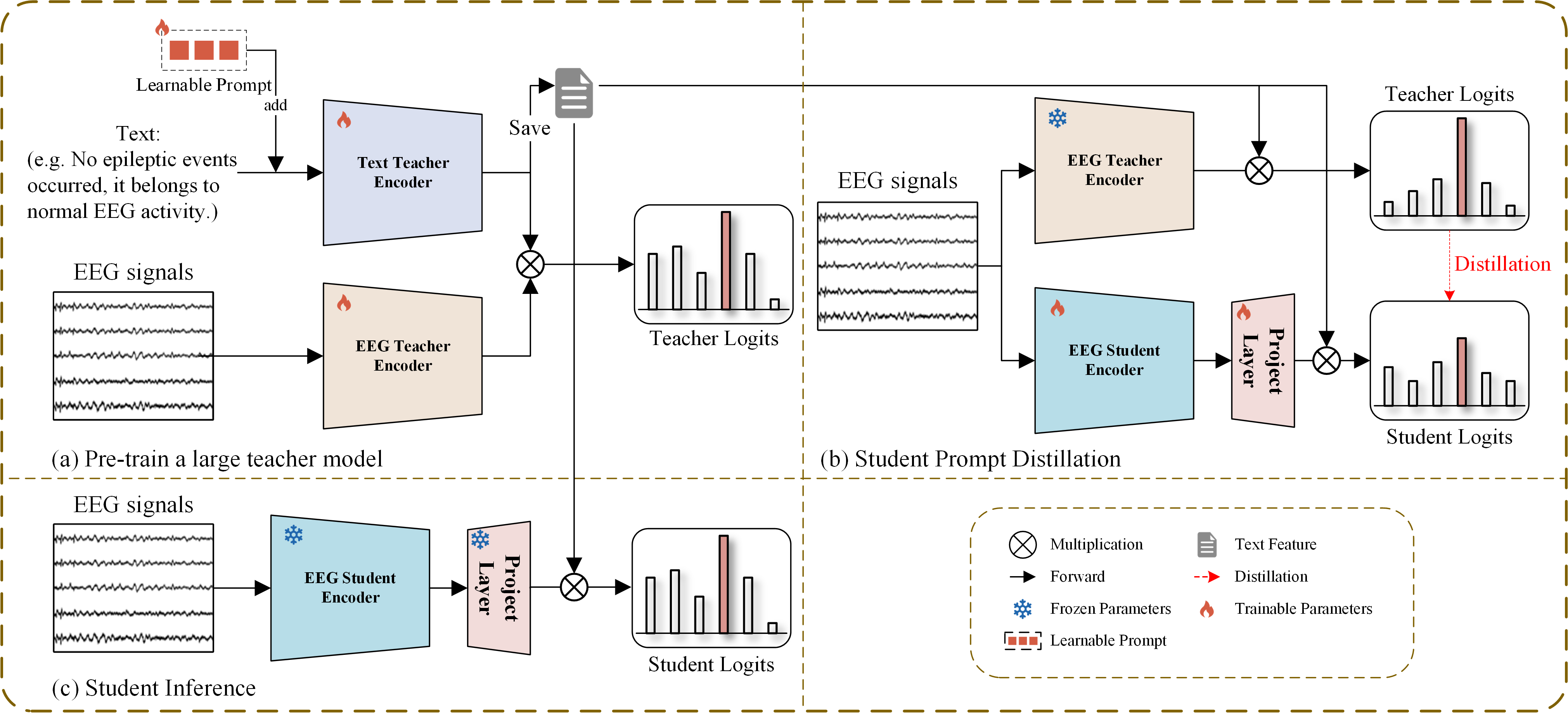}
    \captionsetup{justification=centering}
    \caption{The overall architecture of the proposed DistilCLIP-EEG model.}
    \label{fig:flowchart}
\end{figure*}

Electroencephalography (EEG) is a widely used, non-invasive, and cost-effective tool for real-time monitoring of brain activity, making it ideal for epileptic seizure detection where rapid electrical discharges occur within milliseconds \cite{wong2023eeg}. Compared to traditional imaging methods, EEG offers high temporal resolution and portability, enabling long-term monitoring in both clinical and daily settings. These strengths have driven its broad adoption in medical and research domains.

Artificial intelligence (AI) methods have shown promise in analyzing EEG signals for epileptic seizure detection. While unimodal EEG-based approaches offer high temporal resolution and ease of acquisition, they face notable limitations. EEG signals are often noisy and vulnerable to external interference, hindering robustness across tasks. Furthermore, the lack of contextual information restricts the interpretation of complex brain activities such as emotional states or seizures \cite{EEGAI, seizureEEG}. Consequently, unimodal approaches may suffer from reduced generalizability and suboptimal performance in complex scenarios.

Multimodal learning, which integrates EEG with complementary data such as text descriptions or other biosignals, addresses the limitations of unimodal EEG by enabling richer and more robust representations of brain activity. In neural signal processing, combining EEG with modalities like fNIRS, EMG, and EOG has significantly improved classification accuracy across tasks such as motor imagery, mental workload estimation, and driver vigilance assessment. Deep learning-based multimodal architectures further enhance signal classification and compression with minimal distortion \cite{multimodal_1,multimodal_2}. Notably, EEG-fNIRS fusion improves performance in motor and workload tasks, while EEG-EMG and EEG-EOG combinations yield better results in emotion recognition and vigilance monitoring, respectively \cite{multimodal_4,multimodal_5}.

In the context of epilepsy detection, multimodal methods enhance the model's ability to distinguish between different types of seizures or other neurological events \cite{multimodal_6,multimodal_7}. However, despite their performance advantages, these methods often suffer from high computational costs and increased architectural complexity. These limitations make real-time applications and deployment in resource-constrained settings particularly challenging. Our work addresses these challenges by integrating EEG with text data in a more streamlined and efficient multimodal framework. This design enables accurate and resource-efficient epilepsy detection, maintaining the benefits of multimodal learning while reducing its overhead.

In this work, we propose DistilCLIP-EEG, a novel multimodal framework based on the CLIP architecture \cite{CLIP}, to address the limitations of unimodal EEG-based methods. By integrating EEG signals with text descriptions, the model enhances cross-modal representation learning through a Conformer-based EEG encoder and a BERT-LP text encoder, both operating in a shared latent space. Prompt learning is incorporated to improve contextual understanding and overall performance. To reduce computational complexity, we adopt a knowledge distillation strategy, using the trained CLIP model as a teacher to guide a compact student model. This is the first application of CLIP-based multimodal learning for seizure detection, demonstrating that model distillation enables efficient and accurate performance with reduced overhead. The architecture is illustrated in Fig.~\ref{fig:flowchart}. The main contributions are as follows:

\begin{itemize}
    \item {Multimodal integration}: our method uniquely integrates EEG signals and corresponding text descriptions into a shared latent space. This multimodal fusion enhances the model’s ability to capture the intricate relationships between the two modalities, leading to improved accuracy in seizure detection.
    \item {Knowledge distillation for model compression}: to address computational constraints in resource-limited environments, we introduced a knowledge distillation framework. In this framework, the pre-trained CLIP model and BERT text encoder act as teachers, guiding the training of a more compact student model. By leveraging these powerful pre-trained models, the student model achieves performance close to that of the teacher while significantly reducing computational demands. 
    \item {Learnable prompt integration}: to enhance adaptability and flexibility, we incorporate prompt learning into both the Conformer-based EEG encoder and the BERT-LP text encoder. Instead of relying on manually designed prompts, our approach turns prompts into learnable parameters, which are optimized to extract meaningful representations from downstream datasets, thereby improving generalization across different datasets.
\end{itemize}

These contributions mark a significant advancement in the state-of-the-art for epilepsy detection, providing a robust and efficient solution with strong potential for real-world clinical applications. The remainder of this paper is structured as follows: Section~\ref{sec:Related Work} presents related work. Section~\ref{sec:Methodology} details our proposed approach. Section~\ref{sec:Dataset and Experimental Setup} describes the experimental setup. Section~\ref{sec:Results and Discussion} shows and discusses the experimental results, and Section~\ref{sec:Conclusion} is the conclusion.

\section{Related Work}
\label{sec:Related Work}
Epileptic seizure detection and prediction have emerged as critical challenges in both medical research and engineering. With the rapid progress in deep learning technologies, significant advancements have been made in EEG-based seizure detection methods. In this section, we review relevant research, focusing on EEG-based seizure detection, the application of multimodal learning in medical signal processing, and distillation learning for model compression and optimization.

\subsection{Deep Learning in Epilepsy Detection}
Traditional machine learning methods in epilepsy detection often rely on manual feature extraction and suffer from high computational complexity. In contrast, deep learning integrates feature extraction and classification, offering improved efficiency and performance \cite{Natu}.

Convolutional Neural Networks (CNNs) have demonstrated strong feature extraction capabilities and become a mainstream approach in EEG-based seizure detection \cite{Acharya}. For instance, Shan et al. \cite{Multi-Branch} proposed MultiSincNet, a spatio-temporal multi-branch CNN architecture, which—when combined with ShallowConvNet, DeepConvNet, and EEGWaveNet—achieved notable improvements across various metrics. 

Recurrent Neural Networks (RNNs), particularly LSTM variants, are also widely adopted for temporal modeling. Hussein et al. \cite{Hussein} employed a single-layer LSTM for robust seizure detection under noisy conditions. Tuncer and Bolat \cite{Tuncer} used BiLSTM to classify features like instantaneous frequency and spectral entropy, achieving 99\% accuracy on the Bonn dataset.

Hybrid CNN-RNN models leverage both spatial and temporal features. Asif et al. \cite{Asif} proposed SeizureNet, combining multispectral embeddings and knowledge distillation to enhance performance in compact models. However, issues such as high parameter count, spectrogram-induced information loss, and limited interpretability of 2D convolutions on 1D signals were noted \cite{Wyse}. Ahmad et al. \cite{Ahmad} introduced a 1D CNN-BiLSTM hybrid model that outperformed several baselines on the UCI dataset. 

Despite these advances, most methods remain unimodal, relying solely on EEG signals. Such approaches struggle with complex or noisy data and lack generalization. Moreover, they ignore complementary information from other modalities, such as clinical or textual data. Multimodal methods address these limitations by integrating heterogeneous sources, yielding more robust and generalizable seizure detection models.

\subsection{Multimodal learning in EEG signal processing}
Multimodal learning combines data from different modalities to capture complementary information, thereby improving model accuracy, robustness, and generalizability. In medicine, it has been successfully applied to tasks such as tumor detection, ECG analysis, and medical image classification \cite{Lu}.

In EEG analysis, multimodal learning is still emerging. Some studies have integrated EEG with other biological signals to improve seizure detection. For example, Sirpal et al. \cite{multimodal_6} fused EEG and fNIRS, achieving high sensitivity (89.7\%) and specificity (95.5\%). Hosseini et al. \cite{multimodal_7} combined rs-fMRI with EEG/iEEG and proposed a deep learning-based unsupervised model for preictal state prediction. However, most existing works focus on physiological signal fusion, largely overlooking textual modalities.

The integration of textual descriptions with EEG in multimodal frameworks remains underexplored. To address this gap, we propose a novel CLIP-based approach that jointly embeds EEG signals and their corresponding textual annotations into a shared latent space. This framework demonstrates the potential of leveraging diverse modalities to enhance seizure detection and opens new directions for multimodal EEG research.

\begin{figure*}[!t] 
    \centering
    \includegraphics[width=\textwidth]{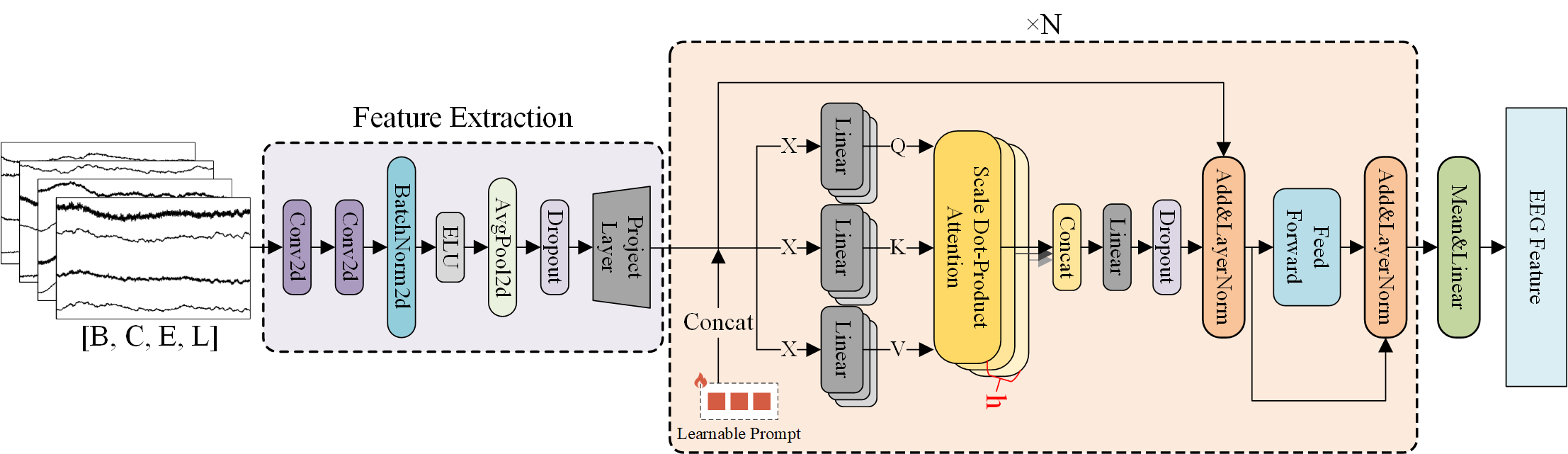}
    \captionsetup{justification=raggedright} 
    \caption{The network architecture of Conformer processes multi-channel EEG 
    signals as input, where B represents the batch size, C denotes the 
    number of channels, E indicates the number of electrodes, and L 
    stands for the number of signal sampling points.}
    \label{fig:Conformer}
  \end{figure*}
\begin{figure}[!t] 
 \centering
 \includegraphics[width=\linewidth]{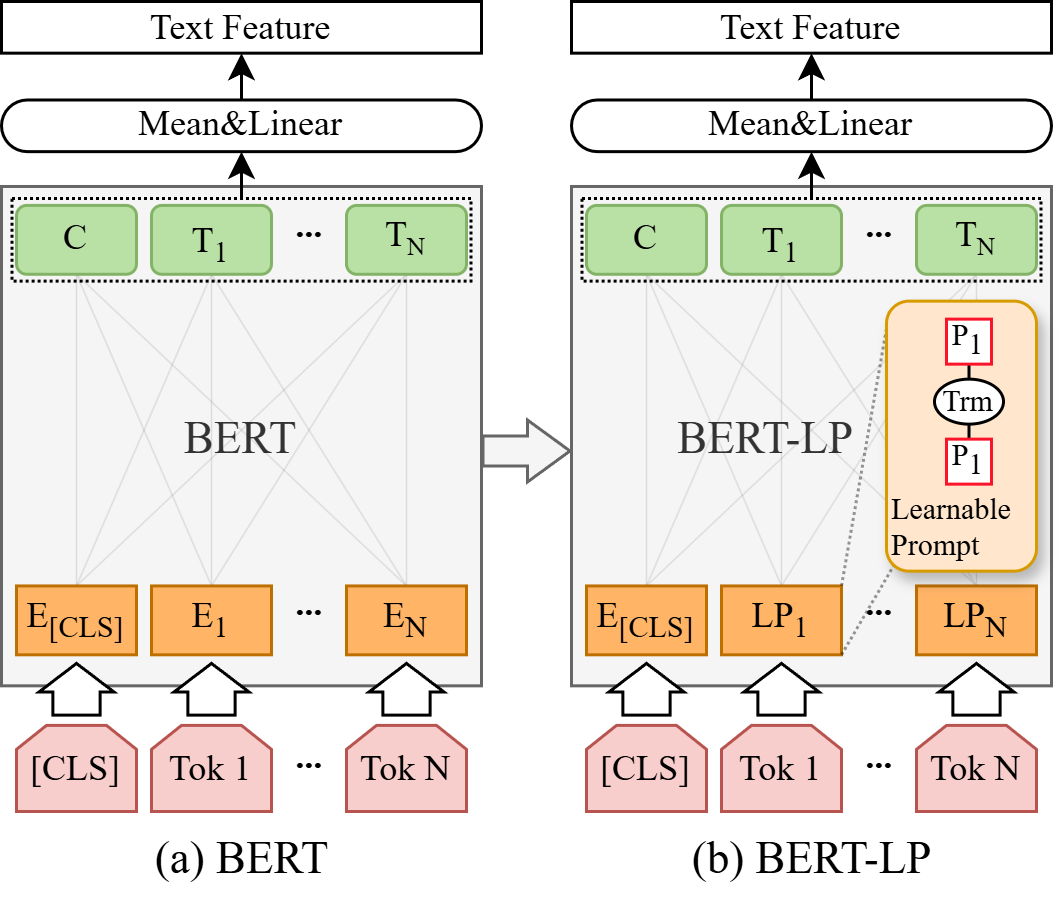}
 \captionsetup{justification=raggedright} 
 \caption{(a) Original BERT; (b) The proposed BERT-LP with learnable prompt}
 \label{fig:Bert-LP}
\end{figure}

\subsection{Distillation learning in deep model compression}
Deep learning models often demand significant computational and storage resources, limiting their deployment in resource-constrained environments. While architectures like CNNs and RNNs achieve high accuracy in domains such as medical imaging and NLP, their complexity hinders practical applications on edge devices. Knowledge distillation addresses this issue by training a compact “student” model to replicate the performance of a larger “teacher” model, thereby reducing resource consumption without substantial loss in accuracy \cite{Hinton}.

This technique has shown success in various fields, including medical image classification, where lightweight models maintain near-teacher performance with reduced inference time and memory usage \cite{Xing}. However, its application in EEG-based epilepsy detection remains underexplored due to the high dimensionality and complexity of EEG signals. To bridge this gap, we introduce a CLIP-based multimodal distillation framework that compresses the model while integrating EEG signals and textual descriptions. Our approach significantly reduces computational overhead while preserving performance, highlighting the potential of knowledge distillation in efficient and scalable medical signal processing.

\section{Methodology}
\label{sec:Methodology}
This section presents a comprehensive description of the architecture and components of the DistilCLIP-EEG model. Designed to improve the accuracy and robustness of epileptic seizure detection, the DistilCLIP-EEG model integrates EEG signals with their corresponding textual descriptions using multimodal learning. The model leverages advanced deep learning techniques, including Conformer, BERT-LP, prompt learning, and knowledge distillation, to address the complexities of EEG signals and the depth of textual semantics.

The DistilCLIP-EEG model processes EEG signals using a Conformer-based encoder, as illustrated in Fig.~\ref{fig:Conformer}. The Conformer architecture, which integrates convolutional operations with self-attention, effectively captures both local features and global temporal dependencies. This capability is essential for addressing the non-stationary and complex characteristics of EEG data.

For the textual modality, the model utilizes BERT-LP, a bidirectional Transformer that encodes structured clinical prompts into high-dimensional semantic representations, as shown in Fig.~\ref{fig:Bert-LP}. These prompts serve as expert-annotated descriptors for each EEG segment and are free of explicit classification labels, thereby preventing label leakage during inference \cite{CLIP, CoOp}. The EEG and textual features are projected into a shared latent space, enabling joint multimodal learning and enhancing feature discriminability under limited supervision.

To improve generalization and adaptability, the model incorporates prompt learning, which dynamically optimizes textual inputs across different datasets and tasks. Furthermore, knowledge distillation is employed to reduce computational cost. As depicted in Fig.~\ref{fig:flowchart}, the student model receives supervision from a larger CLIP-based teacher model by aligning their output logits and reusing the teacher’s high-quality text features. Only the student’s EEG encoder is updated during training, significantly reducing training overhead. This approach preserves much of the teacher model’s accuracy while enabling deployment in resource-limited environments, thereby enhancing the practicality of the DistilCLIP-EEG framework.

We primarily compute the model's cross-entropy loss, which is particularly suited for classification tasks, especially multi-class problems. This loss function combines softmax and log-likelihood to measure the difference between the predicted distribution and the true distribution. Assuming we have \( N \) samples, with each sample belonging to \( C \) classes, the formula for the loss function is:
\begin{equation}
\text{CE} = -\frac{1}{N} \sum_{i=1}^{N} \sum_{j=1}^{C} y_{ij} \log(\hat{y}_{ij})
\label{eq:loss}
\end{equation}
where \( N \) denotes the number of samples, \( C \) represents the total number of classes, and \( y_{ij} \) is the true label of sample \( i \) for class \( j \). The term \( \hat{y}_{ij} \) refers to the predicted probability of sample \( i \) belonging to class \( j \), which is calculated using the softmax function.

\begin{algorithm}[!t]
\caption{Pseudocode for the core of an implementation of DistilCLIP-EEG.}
\label{alg:1}
\begin{algorithmic}
\REQUIRE EEG signal $\mathbf{X_{EEG}} \in \mathbf{D_{train}}$, Text $\mathbf{X_{TXT}} \in \mathbf{D_{train}}$, True label $\mathbf{y} \in \mathbf{D_{train}}$, EEG Encoder $\mathbf{E_{eeg}}$ and Text Encoder $\mathbf{E_{txt}}$
\ENSURE Trained EEG Encoder $\mathbf{E_{eeg}}$ and Text Encoder $\mathbf{E_{txt}}$
\STATE \textbf{Initialize parameters: }Epoch $N$,\,Batch $B$,\,$\theta_{eeg},\,\theta_{txt},\,\alpha$ and $\sigma$
\STATE $x_{eeg} \gets \text{Z-score}(\mathbf{X_{EEG}})$
\STATE $x_{txt} \gets \text{Tokenize}(\mathbf{X_{TXT}})$
\FOR{$n \gets 1$ to $N$}
\FOR{$b \gets 1$ to $B$}
\STATE \textsc{/* Forward pass */}
\STATE $f_{eeg},\:f_{txt} \gets \mathbf{E_{eeg}}(x_{eeg}, \theta_{eeg}),\:\mathbf{E_{txt}}(x_{txt}, \theta_{txt})$
\STATE $V,\:U \gets f_{eeg}\,/\,||f_{eeg}||,\:f_{txt}\,/\,||f_{txt}||$
\STATE $l_{eeg},\:l_{txt} \gets \sigma \cdot V \times U^T,\:\sigma \cdot U \times V^T$
\STATE $L_1 \gets (CE(l_{eeg},\,y)\,+\,CE(l_{txt},\,y))\,/\,2$
\STATE $L_2 \gets CE(\hat{y},\,y)$
\STATE $L_{total} \gets \alpha \cdot L_{1} + (1-\alpha) \cdot L_{2}$
\STATE
\STATE \textsc{/* Backward pass */}
\STATE \textbf{Back-propagate to optimize} $E_{eeg},\,E_{txt},\,\theta_{eeg},\,\theta_{txt},$ and $\sigma$
\ENDFOR
\ENDFOR
\end{algorithmic}
\end{algorithm}

\subsection{Distilclip-EEG}
\label{Distilclip-EEG}
The DistilCLIP-EEG model is a multimodal framework designed for epileptic seizure detection, extending the CLIP architecture to jointly embed EEG signals and corresponding textual descriptions into a shared latent space. It comprises two main components: a Conformer-based EEG encoder and a BERT-LP text encoder, detailed in Sections~\ref{subsec:Conformer} and~\ref{subsec:BERT-LP}. The Conformer captures temporal dependencies in EEG data, while BERT-LP extracts semantic representations from structured clinical prompts. Joint training aligns both modalities, enhancing the model’s ability to leverage complementary information for improved seizure detection. The overall framework is illustrated in Fig.~\ref{fig:flowchart}, with pseudocode provided in Algorithm~\ref{alg:1}.

DistilCLIP-EEG integrates CLIP’s multimodal alignment with Prompt Learning and Knowledge Distillation to improve generalization and model compression. This design not only boosts detection performance but also enables efficient deployment in resource-constrained environments.

\subsection{EEG Encoder: Conformer} 
\label{subsec:Conformer}
Accurate EEG encoding requires a model capable of capturing both local features and global dependencies. We adopt the Conformer architecture, as illustrated in Fig.~\ref{fig:Conformer}, which integrates convolutional layers with Transformer-based self-attention. This hybrid design leverages the strengths of convolutional neural networks for extracting localized spatial and frequency features, while the self-attention mechanism in Transformers effectively models long-range temporal dependencies.

Convolutional layers capture spatial and frequency-localized features, while Transformer layers model long-range temporal dependencies through multi-head self-attention. The Conformer first extracts local representations and then globally contextualizes them, enhancing the modeling of non-stationary EEG signals.

As shown in Fig.~\ref{fig:Conformer}, the multidimensional input structure enables the Conformer model to fully exploit the temporal, spectral, and spatial characteristics of EEG signals. By integrating convolutional layers with self-attention mechanisms, the model effectively captures local patterns and global dependencies across frequency bands, thereby significantly enhancing the accuracy and robustness of feature extraction in complex EEG scenarios.

By combining multi-scale feature extraction with global attention, the Conformer offers robust and accurate representations of EEG signals, effectively handling variability and noise inherent in seizure detection tasks.

\subsection{Text Encoder: BERT-LP} 
\label{subsec:BERT-LP}
In multimodal EEG-text learning, selecting an effective text encoder is essential for capturing semantic information \cite{Devlin}. The DistilCLIP-EEG model employs BERT-LP, a variant of BERT based on the Transformer architecture, to extract contextual representations from EEG-related descriptions. Its bidirectional self-attention and pre-training strategies, including Masked Language Modeling and Next Sentence Prediction, enhance semantic understanding. As illustrated in Fig.~\ref{fig:Bert-LP}, BERT-LP maps textual inputs into high-dimensional embeddings aligned with EEG features. Prompt learning and fine-tuning further enhance feature relevance, facilitating accurate multimodal fusion for seizure detection.

\subsection{Prompt Learning}
This study incorporates Prompt Learning to enhance the performance of both EEG and text encoders. Unlike traditional hand-crafted prompts, which lack adaptability, the DistilCLIP-EEG model treats prompts as learnable parameters optimized during training. This allows dynamic adaptation to downstream data characteristics and improves the quality of extracted features by leveraging domain-specific textual cues.

As shown in Fig.~\ref{fig:Bert-LP}, learnable prompts are inserted into each layer of the BERT-LP Transformer to refine text representations. Similarly, Fig.~\ref{fig:Conformer} illustrates the integration of prompts into Conformer layers, enabling the EEG encoder to adaptively learn task-relevant features without manual design. This approach achieves efficient adaptation with fewer trainable parameters, often surpassing full fine-tuning in performance.

Prompt Learning improves both model adaptability and multimodal fusion. By guiding the representation learning process, it enhances seizure detection accuracy and generalization. It also reduces training complexity by leveraging pre-trained knowledge, allowing effective signal-text alignment across diverse EEG datasets.

\begin{algorithm}[!t]
\caption{Pseudocode for the main DistilCLIP-EEG distillation process.}
\label{alg:2}
\begin{algorithmic}
\REQUIRE Teacher model $\mathbf{T}$, Student model $\mathbf{S}$, train data $x \in \mathbf{D_{train}}$, and true label $y \in \mathbf{D_{train}}$
\ENSURE Trained student models $\hat{S}$
\STATE
\STATE {\textbf{Initialize parameters: }}Epoch $N$,\,Batch $B$, $\alpha$, Student model parameters $\theta_{S}$, and temperature parameter $t$

\FOR{$n \gets 1$ to $N$}
\FOR{$b \gets 1$ to $B$}
\STATE \textsc{/* Forward pass */}
\STATE $\hat{y}_T,\:\hat{y}_S \gets \mathbf{T}(x),\:\mathbf{S}(x)$
\STATE $\hat{y}_T^T,\:\hat{y}_S^T \gets \text{softmax}(\hat{y}_T\,/\,t),\:\text{softmax}(\hat{y}_S\,/\,t)$
\STATE $L_{KL} \gets KL(\hat{y}_S^T,\,\hat{y}_T^T)$
\STATE $L_{CE} \gets CE(\hat{y}_S,\,y)$
\STATE $L_{total} \gets \alpha \cdot L_{CE} + (1 - \alpha) \cdot t^2 \cdot L_{KL}$
\STATE
\STATE \textsc{/* Backward pass */}
\STATE \textbf{Update} $\theta_S$ by $L_{total}$
\ENDFOR
\ENDFOR
\end{algorithmic}
\end{algorithm}

\subsection{Knowledge Distillation}
Knowledge Distillation is a model compression strategy that transfers the knowledge of a large, high-performing teacher model to a smaller, more efficient student model. This approach reduces computational costs while preserving model performance, making it well-suited for resource-intensive deep learning tasks.

In the DistilCLIP-EEG framework, a pre-trained CLIP model serves as the teacher, guiding the training of a lightweight student model. The student learns to mimic the teacher’s output distribution, with Kullback-Leibler (KL) divergence used to minimize the discrepancy between their predictions:
\begin{equation}
\text{KL}(P \parallel Q) = \frac{1}{N} \sum_{n=1}^{N} \sum_{i=1}^{C} Q_i^{(n)} \log \frac{Q_i^{(n)}}{P_i^{(n)}}
\label{eq:kl}
\end{equation}
\begin{equation}
P_i^{(n)} = \frac{\exp(s_i^{(n)})}{\sum_{j=1}^{C} \exp(s_j^{(n)})}
\label{eq:P}
\end{equation}
\begin{equation}
Q_i^{(n)} = \frac{\exp(t_i^{(n)})}{\sum_{j=1}^{C} \exp(t_j^{(n)})}
\label{eq:Q}
\end{equation}
where \( N \) is the number of samples and \( C \) is the number of classes. Equation (\ref{eq:P}) represents the target distribution or true distribution, while (\ref{eq:Q}) represents the model's predicted distribution.

This distillation process enables the student to replicate the teacher’s decision-making ability with significantly fewer parameters. In EEG-based seizure detection, this results in models that are both accurate and deployable in resource-constrained settings. The procedure is outlined in Algorithm~\ref{alg:2}.

By compressing a multimodal CLIP-based model through Knowledge Distillation, DistilCLIP-EEG achieves a strong balance between performance and efficiency, expanding the feasibility of multimodal learning in practical clinical applications.

\section{Dataset and Experimental Setup} 
\label{sec:Dataset and Experimental Setup}
\subsection{Dataset}
\begin{figure}[!t] 
    \centering
    \includegraphics[width=0.65\linewidth]{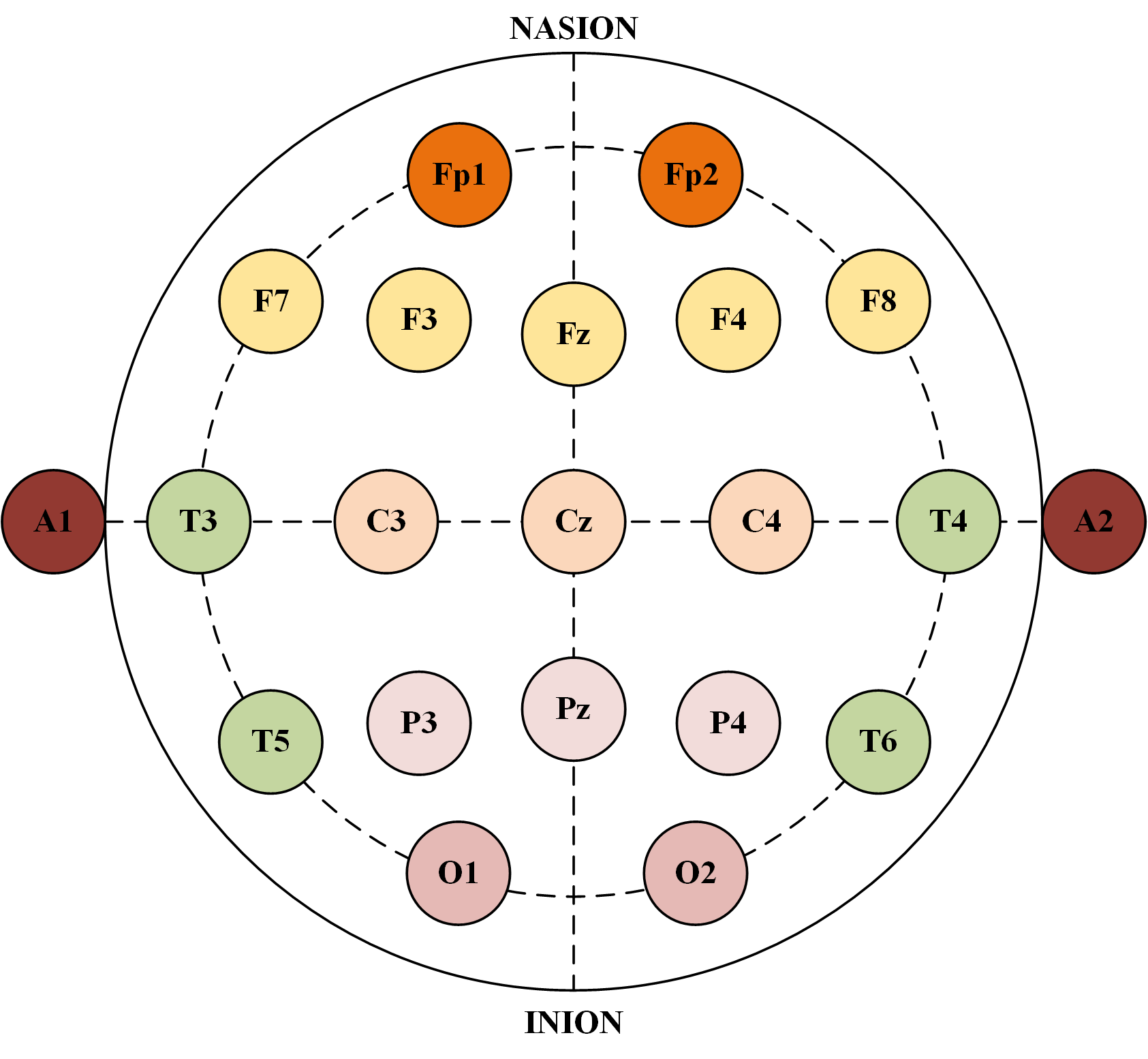}
    \captionsetup{justification=raggedright} 
    \caption{Electrode distribution map of the international 10-20 system for EEG electrode placement.}
    \label{fig:10-20}
\end{figure}

\subsubsection{TUSZ}
This study utilizes the Temple University Hospital EEG (TUSZ v2.0.1) dataset, which contains EEG recordings from 675 patients. Among them, over 280 patients experienced at least one epileptic seizure, contributing to more than 3,500 recorded seizure events. The total recording duration exceeds 5 million seconds, with seizure segments accounting for approximately 4–6\% of the data. 

TUSZ is one of the largest and most diverse publicly available EEG datasets for epilepsy research, offering rich clinical metadata, including patient age, gender, medical history, seizure frequency, and medication usage \cite{TUSZ, Shah}. It covers eight seizure types: FNSZ, GNSZ, ABSZ, CPSZ, TCSZ, TNSZ, SPSZ, and MYSZ, with BCKG annotated as background activity. Due to the rarity of MYSZ events, this category was excluded from our analysis, which focused on the remaining seven seizure types.

\subsubsection{AUBMC}
The AUBMC dataset comprises EEG recordings from six patients with focal epilepsy, collected at the American University of Beirut Medical Center during long-term video-EEG monitoring for pre-surgical evaluation \cite{AUBMC}. To capture habitual seizures, anti-epileptic medications were withdrawn during recording. A total of 20 EEG sessions were acquired, each containing interictal segments and one or more ictal events. Signals were recorded using 19 scalp electrodes according to the 10–20 system at a 500 Hz sampling rate, as illustrated in Fig.~\ref{fig:10-20}. Each recording was annotated with seizure type, ictal onset zone, and seizure duration, with total ictal activity exceeding one hour. The data is categorized into four classes: normal, complex partial seizures, electrographic seizures, and seizures without visible electrographic changes detected by video-EEG. This dataset provides a valuable benchmark for assessing model generalization in small-sample, real-world clinical settings.

\subsubsection{CHB-MIT}
The CHB-MIT Scalp EEG dataset, collected at Boston Children’s Hospital, comprises long-term EEG recordings from pediatric patients with intractable epilepsy. To capture habitual seizures, anti-epileptic medications were withdrawn during continuous monitoring over several days as part of pre-surgical evaluation \cite{chbmit, CHB-MIT}.

The dataset includes recordings from 23 cases representing 22 subjects (5 males aged 3–22 years and 17 females aged 1.5–19 years). Case chb21 is a follow-up of chb01, recorded 1.5 years later. Case chb24, added in 2010, extends the original dataset. Each case contains 9–42 recordings (.edf format), most approximately one hour in length; exceptions include chb10 (2-hour files) and chb04, chb06, chb07, chb09, and chb23 (4-hour files).

EEG signals were sampled at 256 Hz with 16-bit resolution using 23 scalp electrodes based on the 10–20 system (some files include 24 or 26 channels). In total, the dataset comprises 664 recordings, with 129 containing seizures, accounting for 198 seizure events. It spans approximately 967.85 hours of EEG data .

\subsubsection{Zenodo}
The Zenodo dataset comprises multi-channel EEG recordings from 79 term neonates admitted to the NICU at Helsinki University Hospital, with a median recording duration of 74 minutes. Three independent experts visually reviewed the recordings, each annotating approximately 460 seizures. Based on consensus, 39 neonates were identified as having seizures, 22 as seizure-free, and the remainder as inconclusive \cite{Zenodo}.

To standardize preprocessing, all recordings were segmented into 1-second epochs. For non-seizure periods, non-overlapping segments were used, while seizure periods were segmented using overlapping windows to better capture the short and sparse nature of seizure events. Specifically, 50\% overlap was applied to seizure segments in the CHB-MIT and Zenodo datasets. For the TUSZ and AUBMC datasets, a 75\% overlap was used to account for the lower prevalence of certain seizure types. Epochs overlapping with any part of a seizure were labeled as seizure, and others as non-seizure.To address the inherent class imbalance issue in seizure detection tasks, we constructed the training dataset by retaining all seizure segments and randomly sampling a similar number of non-seizure segments. This approach facilitates more effective learning of discriminative features for seizure detection across diverse clinical datasets.

\subsection{Preprocessing}
EEG recordings typically contain both seizure and background activity, requiring standardized preprocessing to ensure consistent analysis across datasets. All signals were band-pass filtered between 0.1–70 Hz and underwent artifact removal to enhance data quality.

For the TUSZ dataset, we selected 19 common channels following the international 10–20 system and resampled all signals to 256 Hz. The AUBMC dataset, preprocessed by the data providers, retained its original 500 Hz sampling rate and default channel configuration. CHB-MIT recordings were resampled to 256 Hz, with 23 scalp electrodes selected according to the 10–20 system. Similarly, Zenodo recordings were resampled to 256 Hz and 19 standard channels were selected according to the 10-20 system, as illustrated in Fig.~\ref{fig:10-20}.

These preprocessing procedures standardize input across datasets, ensuring fair and robust evaluation of the proposed model in diverse clinical contexts.

After band-pass filtering, all EEG signals were standardized using Z-score normalization to reduce inter-subject and inter-session variability. This transformation ensures consistent amplitude scaling, facilitating stable model training by converting each channel's signal to a zero-mean, unit-variance distribution. To extract seizure segments, EEG recordings were sliced based on annotated seizure onset and offset times, adjusted according to the sampling frequency. This segmentation, combined with resampling, ensured consistent input representation across datasets, enabling robust model training.

\subsection{Experimental Setup}
In this study, to ensure robust model evaluation, we employed five-fold cross-validation across all four datasets. Each dataset was randomly divided into five non-overlapping subsets, with four used for training and one for testing in each fold. This process was repeated five times, and the final results were averaged. Strict separation between training and testing data was maintained to avoid data leakage. All models were evaluated under consistent training settings to ensure comparability.

As shown in Fig.~\ref{fig:flowchart}, the teacher model was trained for 400 epochs using an 8-layer Conformer encoder, where each block included a 10-head multi-head attention mechanism and a feedforward network with a 4$\times$ expansion ratio. Residual connections and a dropout rate of 0.3 were applied to both attention and feedforward layers to enhance generalization and stability. The output and input embedding dimensions were kept equal.

The BERT-LP module employed a modified 12-layer BERT architecture with bidirectional Transformer encoders, each having 12 attention heads and a hidden size of 768. A prompt mechanism was inserted at each layer to enrich contextual representation, with dropout rates set to 0.2 for both attention and hidden layers.

The student model, a 4-layer Conformer encoder aligned with the teacher architecture, was trained for 100 epochs. A projection layer was included to match the embedding space of the teacher. With approximately 17.9 million parameters—42\% fewer than the teacher model's 30.8 million—the student model achieves substantial reductions in GPU memory usage and inference time.

To ensure optimal model performance, we conducted hyperparameter tuning using a combination of grid search and manual refinement. The tuning process focused on key parameters such as learning rate, batch size, dropout ratio, and optimization strategy. The learning rate was explored in the range from $1\times10^{-5}$ to $1\times10^{-3}$, with $1\times10^{-5}$ ultimately selected as the optimal value based on validation performance. Batch sizes of 16, 32, and 64 were tested, and a batch size of 32 was chosen as a balance between training efficiency and model generalization. Dropout rates were systematically adjusted between 0.1 and 0.5 to prevent overfitting, with final dropout settings of 0.3 in the Conformer encoder and 0.2 in the BERT-LP module. The Adam optimizer was employed with weight decay, and its momentum parameters were set to ($\beta_1=0.9, \beta_2=0.999$). 

\begin{table*}[!t]
  \centering
  \caption{Comparison of our proposed DistilCLIP-EEG model with existing methods on four benchmark EEG datasets}
  \label{tab:related work}
  \resizebox{\textwidth}{!}{ %
  \begin{tabular}{lcccccccccccccccccccc}
    \toprule
    \multirow{2}{*}{Method} & 
    \multicolumn{4}{c}{TUSZ} & 
    \multicolumn{4}{c}{AUBMC} & 
    \multicolumn{4}{c}{CHB-MIT} & 
    \multicolumn{4}{c}{Zenodo} \\
    \cmidrule(lr){2-5} \cmidrule(lr){6-9} \cmidrule(lr){10-13} \cmidrule(lr){14-17} \cmidrule(lr){18-21}
    & Acc(\%) & Pre(\%) & Rec(\%) & F1(\%) 
    & Acc(\%) & Pre(\%) & Rec(\%) & F1(\%) 
    & Acc(\%) & Pre(\%) & Rec(\%) & F1(\%)
    & Acc(\%) & Pre(\%) & Rec(\%) & F1(\%) \\
    \midrule
    CNN-RNN \cite{Tennison}, 2020  
        & 90.61 & 91.45 & 85.11 & 90.86 
        & 91.08 & 93.68 & 87.92 & 92.05 
        & 80.13 & 82.90 & 75.92 & 79.26
        & 81.84 & 83.08 & 79.95 & 81.49
        \\
    GASF-CNN \cite{Shankar}, 2021   
        & 89.56 & 90.67 & 84.39 & 89.86 
        & 88.38 & 91.39 & 81.69 & 89.57 
        & 85.29 & 87.15 & 82.80 & 84.92 
        & 84.03 & 83.82 & 84.33 & 84.07
        \\
    SincNet-Conv1D  \cite{Priyasad}, 2021   
        & 90.05 & 90.92 & 85.04 & 90.24 
        & 89.67 & 92.69 & 85.02 & 90.77 
        & 87.01 & 87.60 & 86.24 & 86.91 
        & 84.42 & 84.22 & 84.73 & 84.47
        \\
    GGN \cite{Li}, 2022             
        & 94.41 & 94.42 & 90.32 & 94.38 
        & 92.23 & 93.82 & 86.31 & 92.85 
        & 92.17 & 94.39 & 89.68 & 91.97 
        & 87.21 & 85.31 & 89.90 & 87.54
        \\
    LightSeizureNet \cite{LightSeizureNet}, 2022
        & 88.64 & 89.85 & 84.29 & 88.65 
        & 90.18 & 92.99 & 85.34 & 91.26 
        & \underline{97.09} & 96.53 & 97.69 & \underline{97.11} 
        & 91.52 & 91.58 & 91.44 & 91.51 
        \\
    EEGNet \cite{Zhu}, 2023         
        & 90.90 & 91.60 & 87.86 & 91.04 
        & 90.95 & 92.47 & 85.48 & 91.54 
        & 89.59 & 86.98 & 93.12 & 89.95 
        & 87.61 & 85.96 & 89.90 & 87.88
        \\
    Meta-GNN \cite{Rahmani}, 2023   
        & 91.46 & 92.22 & 88.57 & 91.61 
        & 88.38 & 91.11 & 81.33 & 89.47 
        & 89.56 & 88.33 & 91.16 & 89.72 
        & 83.62 & 82.16 & 85.92 & 84.00
        \\
    3D-CBAMNet  \cite{Huang}, 2023  
        & \underline{97.00} & \underline{96.98} 
        & \underline{93.11} & \underline{96.98} 
        & \underline{97.66} & \underline{97.75}
        & \underline{91.84} & \underline{97.70} 
        & 96.87 & \underline{97.39} & 96.32 & 96.85
        & \underline{93.57} & 93.31 
        & \underline{93.87} & \underline{93.59}
        \\
    RNN \cite{Statsenko}, 2023      
        & 89.35 & 90.48 & 84.50 & 89.79 
        & 90.99 & 93.16 & 87.60 & 91.81 
        & 86.52 & 86.89 & 86.00 & 86.45 
        & 83.83 & 83.76 & 83.93 & 83.84
        \\

    CNN and RNNs \cite{Yang}, 2024  
        & 79.80 & 83.13 & 78.79 & 80.94 
        & 79.44 & 84.87 & 78.58 & 81.34 
        & 77.91 & 75.54 & 82.56 & 78.90
        & 71.90 & 75.97 & 64.04 & 69.50
        \\
    ResBiLSTM \cite{Zhao}, 2024     
        & 91.53 & 92.34 & 87.11 & 91.81 
        & 92.27 & 93.77 & 88.23 & 92.85 
        & 88.24 & 83.92 & 94.60 & 88.94 
        & 83.82 & 81.28 & 87.91 & 84.46
        \\
    Stacked LSTM \cite{Pandey}, 2024
        & 96.93 & 96.95 & 92.74 & 96.93 
        & 96.12 & 96.68 & 90.53 & 96.32
        & 92.17 & 96.06 & 87.96 & 91.82
        & 90.79 & 91.53 & 89.90 & 90.71
        \\
    PCC-CNN \cite{Shi}, 2024        
        & 91.04 & 91.75 & 86.33 & 91.09 
        & 91.89 & 93.29 & 88.04 & 92.40 
        & 89.59 & 90.93 & 87.96 & 89.42
        & 90.99 & 90.90 & 91.09 & 91.00
        \\
    Two-Layer LSTM \cite{two-layerLSTM}, 2024
        & 82.28 & 84.85 & 80.12 & 82.80
        & 82.48 & 86.88 & 73.05 & 84.26
        & 82.47 & 81.55 & 83.94 & 82.73
        & 79.58 & 79.63 & 79.51 & 79.57
        \\
    GCN-BiGRU \cite{xu}, 2024
        & 91.27 & 91.83 & 86.17 & 91.41 
        & 91.46 & 94.22 & 85.98 & 92.58 
        & 97.05 & 95.92 & \underline{98.28} & 97.09 
        & 93.50 & \underline{95.37} & 91.44 & 93.37 
        \\
    MultiSincNet \cite{MultiSincNet}, 2024
        & 86.49 & 88.63 & 83.00 & 86.87 
        & 87.09 & 92.37 & 83.49 & 89.11 
        & 89.31 & 89.02 & 89.68 & 89.35 
        & 85.87 & 86.21 & 85.41 & 85.81 
        \\
    
    \midrule
    DistilCLIP-EEG Teacher (Ours)  
        & \textbf{97.75} & \textbf{97.82} & \textbf{96.26} & \textbf{97.77} 
        & \textbf{97.77} & \textbf{97.95} & \textbf{96.27} & \textbf{97.82} 
        & \textbf{97.90} & \textbf{98.13} & \textbf{97.66} & \textbf{97.89} 
        & \textbf{94.61} & \textbf{95.78} & \textbf{93.33} & \textbf{94.54} 
    \\
    DistilCLIP-EEG Student (Ours)  
        & \textbf{97.12} & \textbf{97.21} & \textbf{95.86} & \textbf{97.14} 
        & \textbf{97.38} & \textbf{97.64} & \textbf{95.77} & \textbf{97.46} 
        & \textbf{97.13} & \textbf{97.90} & \textbf{96.32} & \textbf{97.10} 
        & \textbf{94.10} & \textbf{95.00} & \textbf{93.10} & \textbf{94.04} 
        \\
    \bottomrule
  \end{tabular}
}
\end{table*}
\begin{figure*}[!t] 
    \centering
    \includegraphics[width=1\linewidth]{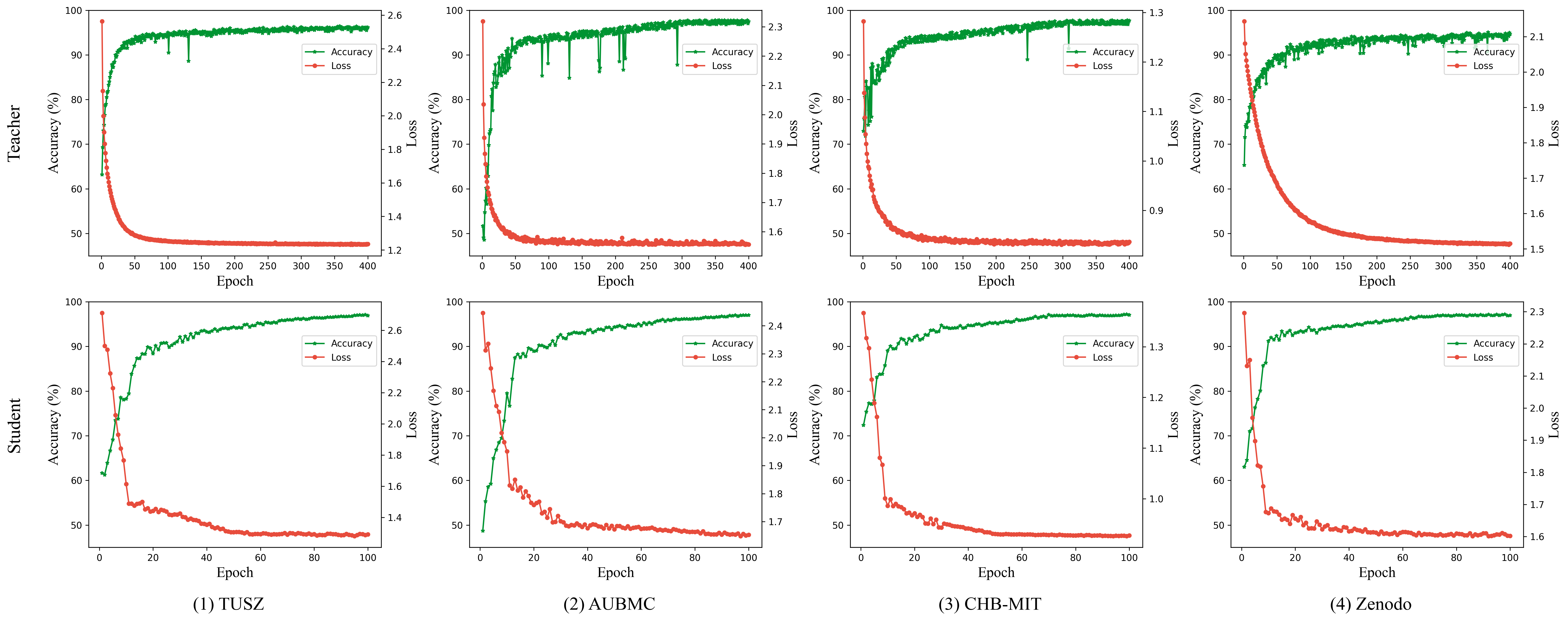}
    \caption{Training accuracy and loss curves of the teacher and student models on four EEG datasets.}
    \label{fig:acc-loss}
\end{figure*}

We also tuned the temperature parameter $t$ in knowledge distillation, evaluating values in {1, 3, 5}. A value of $t=1$ yielded the best performance by balancing knowledge transfer and training stability. All hyperparameters were selected based on development set performance, using F1 score as the selection criterion. Training was implemented in PyTorch on NVIDIA RTX 3090 GPUs with mixed-precision enabled.

\section{Results and Discussion} 
\label{sec:Results and Discussion}
This section compares the proposed DistilCLIP-EEG model with recent state-of-the-art approaches for epileptic seizure detection in terms of methodology, architecture, datasets, and key metrics including accuracy, F1 score, and specificity. As summarized in Table~\ref{tab:related work}, our model consistently outperforms prior methods, particularly in accuracy and F1 score, owing to its multimodal integration, prompt-based learning, and knowledge distillation strategies that enhance feature representation and generalization. Unlike earlier models that rely solely on EEG signals or deep networks such as CNNs and RNNs, DistilCLIP-EEG jointly encodes EEG data and textual context, enabling robust seizure pattern learning. The Conformer EEG encoder and BERT-LP text encoder effectively capture temporal and semantic features, while learnable prompts adaptively guide feature extraction across datasets.

\begin{figure}[!t] 
    \centering
    \includegraphics[width=\linewidth]{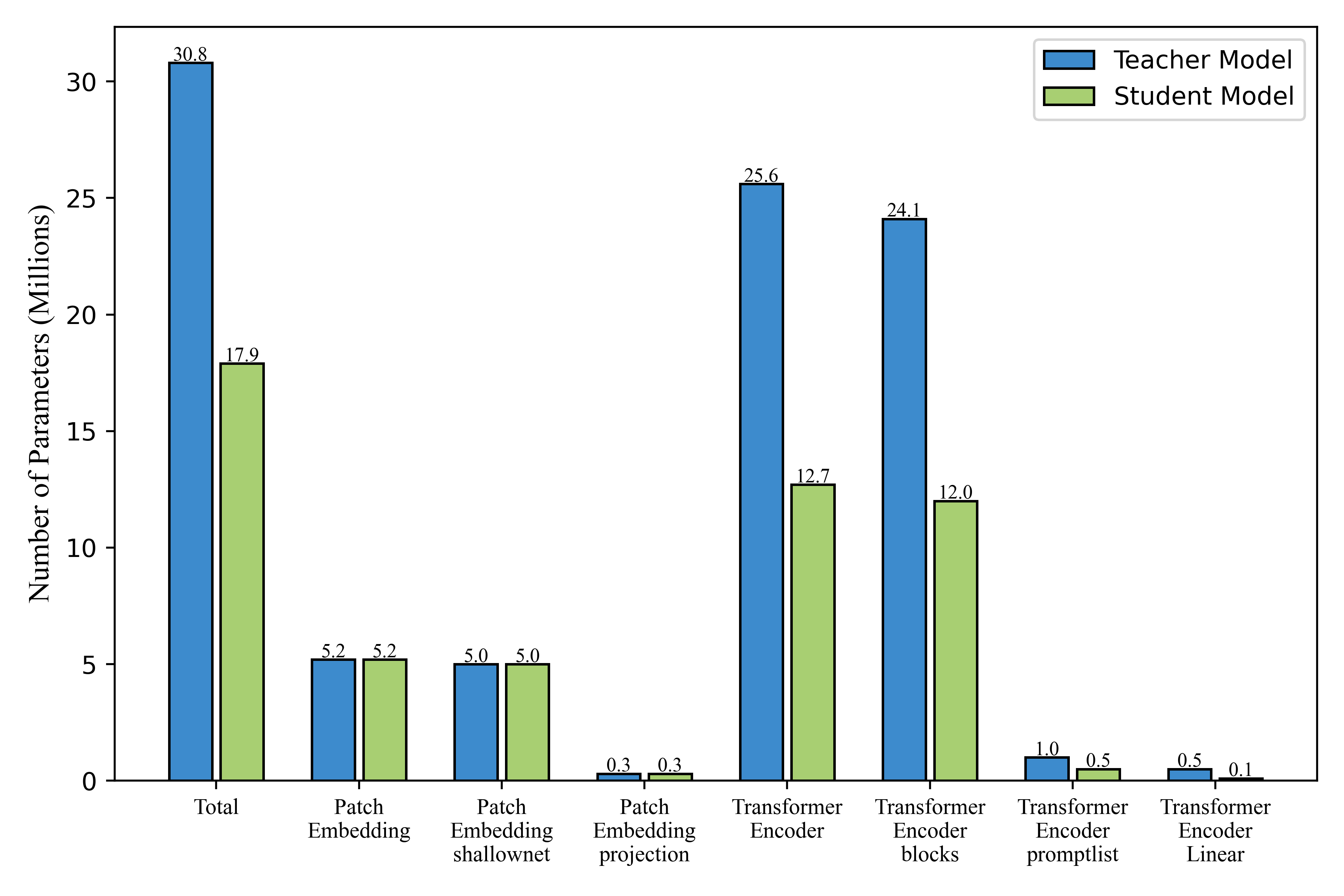}
    \caption{The comparison of the parameter count between the teacher and student model EEG encoders.}
    \label{fig:param}
\end{figure}
\begin{table*}[!t]
  \centering
  \caption{Ablation study results of DistilCLIP-EEG on four benchmark EEG datasets}
  \label{tab:ablation}
  \resizebox{\textwidth}{!}{
  \begin{tabular}{lccccccccccccccccccccc}
    \toprule
    \multirow{2}{*}{Method} & \multicolumn{5}{c}{TUSZ} & \multicolumn{5}{c}{AUBMC} 
    & \multicolumn{5}{c}{CHB-MIT} & \multicolumn{5}{c}{Zenodo}\\
    \cmidrule(lr){2-6} \cmidrule(lr){7-11} \cmidrule(lr){12-16} \cmidrule(lr){17-21}
    & Acc(\%) & Pre(\%) & Rec(\%) & Spe(\%) & F1(\%) 
    & Acc(\%) & Pre(\%) & Rec(\%) & Spe(\%) & F1(\%) 
    & Acc(\%) & Pre(\%) & Rec(\%) & Spe(\%) & F1(\%)
    & Acc(\%) & Pre(\%) & Rec(\%) & Spe(\%) & F1(\%)
    \\
    \midrule
    EEG encoder with learnable prompt (EEG-LP)   
    & 97.37 & 97.42 & 96.25 & 99.62 & 97.38   
    & 97.05 & 97.37 & 94.04 & 98.98 & 97.16 
    & 97.10 & 97.11 & 97.09 & 97.11 & 97.10
    & 94.08 & 95.22 & 92.81 & 95.34 & 94.00
    \\
    Only Text encoder without prompt (Text-WP)  
    & 97.09 & 97.14 & 95.76 & 99.58 & 97.11   
    & 96.92 & 97.27 & 93.44 & 98.95 & 97.05
    & 97.07 & 97.08 & 97.06 & 97.08 & 97.07 
    & 93.96 & 95.01 & 92.80 & 95.12 & 93.89
    \\
    Text encoder with learnable prompt (Text-LP)
    & 96.94 & 96.99 & 95.62 & 99.55 & 96.96   
    & 96.79 & 97.21 & 94.09 & 98.91 & 96.93
    & 97.02 & 96.95 & 97.08 & 96.96 & 97.02
    & 93.82 & 94.91 & 92.60 & 95.03 & 93.74
    \\
    Text encoder with handcrafted prompt (Text-HP)
    & 96.86 & 96.91 & 95.64 & 99.54 & 96.88   
    & 96.66 & 97.11 & 93.52 & 98.86 & 96.82
    & 96.90 & 96.79 & 97.02 & 96.78 & 96.91
    & 93.73 & 94.83 & 92.51 & 94.96 & 93.65
    \\
    Basemodel without prompt (Base-WP)
    & 96.76 & 96.82 & 95.64 & 99.53 & 96.78   
    & 96.53 & 97.00 & 92.89 & 98.82 & 96.70
    & 96.84 & 96.66 & 97.04 & 96.64 & 96.85
    & 93.61 & 94.55 & 92.55 & 94.66 & 93.54
    \\
    \midrule
    Basemodel with learnable prompt (Base-LP)      
    & 97.75 & 97.82 & 96.26 & 99.68 & 97.77   
    & 97.77 & 97.95 & 96.27 & 99.23 & 97.82 
    & 97.90 & 98.13 & 97.66 & 98.14 & 97.90
    & 94.61 & 95.78 & 93.33 & 95.89 & 94.54
    \\
    \bottomrule
  \end{tabular}
}
\end{table*}

As shown in Table~\ref{tab:related work}, both the teacher and student models achieve over 94\% accuracy across all datasets, exceeding 97\% on TUSZ, AUBMC, and CHB-MIT. Despite its reduced complexity, the student model closely matches the teacher in both accuracy and F1 score, demonstrating the effectiveness of the knowledge distillation framework. This performance-efficiency trade-off makes the student model particularly suitable for deployment in resource-constrained clinical environments. In the table, bold values indicate the best results, and underlined values indicate the second-best, further highlighting the superior performance of our proposed models.

\begin{table}[!t]
    \centering
    \caption{The results of the TUSZ dataset, including the teacher model and the student model.}
    \label{TUSZ results}
    \resizebox{\linewidth}{!}{
    \begin{tabular}{c c c c c c}
    \toprule
    \textbf{Class} & \textbf{Acc(\%)} & \textbf{Pre(\%)} & \textbf{Recall(\%)} & \textbf{Spec(\%)} & \textbf{F1(\%)} \\ 
    \midrule
    ABSZ & 85.17/85.17 & 75.11/75.11 & 85.17/85.17   & 99.58/99.58 & 79.82/79.82 \\
    BCKG & 97.70/97.70 & 99.25/99.25 & 97.70/97.70   & 99.80/99.80 & 98.47/98.47 \\
    CPSZ & 97.82/97.34 & 98.15/98.15 & 97.82/97.34   & 99.69/99.69 & 97.99/97.74 \\
    FNSZ & 97.10/94.43 & 99.28/98.92 & 97.10/94.43   & 99.81/99.72 & 98.18/96.62 \\
    GNSZ & 99.27/99.27 & 98.25/95.72 & 99.27/99.27   & 99.53/98.82 & 98.76/97.46 \\
    SPSZ & 97.50/97.50 & 95.59/95.59 & 97.50/97.50   & 99.66/99.66 & 96.53/96.53 \\
    TCSZ & 98.14/98.14 & 96.61/96.61 & 98.14/98.14   & 99.63/99.63 & 97.37/97.37 \\
    TNSZ & 97.39/97.39 & 94.28/94.28 & 97.39/97.39   & 99.75/99.75 & 95.81/95.81 \\
    \midrule
    Overall  & 97.75/97.12 & 97.82/97.21 & 96.26/95.86 & 99.68/99.58 & 97.77/97.14 \\
    \bottomrule
    \end{tabular}
    }
  \end{table} 
\begin{table}[!t]
  \centering
  \caption{The results of the AUBMC dataset, including the teacher model and the student model.}
  \label{AUBMC results}
  \resizebox{\linewidth}{!}{
  \begin{tabular}{c c c c c c}
  \toprule
  \textbf{Class} & \textbf{Acc(\%)} & \textbf{Pre(\%)} & \textbf{Recall(\%)} & \textbf{Spec(\%)} & \textbf{F1(\%)} \\
  \midrule
  \makecell{Normal} 
  & 97.33/97.33 & 99.48/99.14 & 97.33/97.33 & 99.49/99.15 & 98.39/98.23 \\
  \makecell{Complex Partial Seizures} 
  & 98.95/98.29 & 97.91/97.83 & 98.95/98.29 & 98.65/98.61 & 98.43/98.06 \\
  \makecell{Electrographic Seizures} 
  & 96.00/94.67 & 94.61/94.29 & 96.00/94.67 & 99.42/99.39 & 95.30/94.48 \\
  \makecell{Video-detected Seizures\\with no visual change over EEG} 
  & 92.79/92.79 & 67.76/62.42 & 92.79/92.79 & 99.36/99.19 & 78.33/74.64 \\
  \midrule
  \makecell{Overall} 
  & 97.77/97.38 & 97.95/97.64 & 96.27/95.77 & 99.23/99.09 & 97.82/97.46 \\
  \bottomrule
  \end{tabular}
  }
\end{table}
\begin{table}[!t]
    \centering
    \caption{Area under the curve (AUC) scores of the teacher and student models across four EEG benchmark datasets}
    \label{tab:AUC}
    \resizebox{\linewidth}{!}{
    \begin{tabular}{c c c c c}
      \toprule
      \textbf{Dataset} & \textbf{Class} & \textbf{AUC} & \textbf{AUC of macro} & \textbf{AUC of micro} \\ 
      \midrule
      \multirow{8}{*}{\centering TUSZ} 
      & ABSZ & 0.9237/0.9237 
      & \multirow{8}{*}{\makecell{0.9797 /\\ 0.9772}} 
      & \multirow{8}{*}{\makecell{0.9872 /\\ 0.9836}} \\ 
      & BCKG & 0.9875/0.9875 & \\
      & CPSZ & 0.9875/0.9851 & \\
      & FNSZ & 0.9846/0.9708 & \\
      & GNSZ & 0.9940/0.9904 & \\
      & SPSZ & 0.9858/0.9858 & \\
      & TCSZ & 0.9888/0.9888 & \\
      & TNSZ & 0.9857/0.9857 & \\
      \cmidrule{1-5}
      \multirow{5}{*}{\centering AUBMC} 
      & Normal & 0.9841/0.9824 
      & \multirow{5}{*}{\makecell{0.9775 /\\ 0.9743}} 
      & \multirow{5}{*}{\makecell{0.9851 /\\ 0.9825}} \\ 
      & Complex Partial Seizures & 0.9880/0.9845 &  \\ 
      & Electrographic Seizures & 0.9771/0.9703 &  \\ 
      & \makecell{Video-detected Seizures\\with no visual change over EEG} & 0.9608/0.9599 &  \\
      \cmidrule{1-5}
      \multirow{2}{*}{\centering CHB-MIT} 
      & Normal & \multirow{2}{*}{\makecell{0.9790/0.9713}}
      & \multirow{2}{*}{\makecell{-}} 
      & \multirow{2}{*}{\makecell{-}} \\
      & Seizure & \\
      \cmidrule{1-5}
      \multirow{2}{*}{\centering Zenodo} 
      & Normal & \multirow{2}{*}{\makecell{0.9461/0.9410}}
      & \multirow{2}{*}{\makecell{-}} 
      & \multirow{2}{*}{\makecell{-}} \\
      & Seizure & \\
      \bottomrule
    \end{tabular}
    }
  \end{table}

Fig.~\ref{fig:acc-loss} illustrates training dynamics of both models across the four benchmark datasets. The teacher model (top row) exhibits stable convergence and reaches over 95\% accuracy with minimal loss after 300–400 epochs. The student model (bottom row), trained via distillation, converges more quickly and achieves comparable accuracy with only a slight increase in final loss, highlighting its training efficiency.

The lightweight student model effectively preserves the key discriminative knowledge from the teacher model, making it well-suited for deployment in resource-constrained environments. As shown in Fig.~\ref{fig:param}, the teacher model’s EEG encoder comprises approximately 30.8 million parameters, while the student model contains only 17.9 million—a 42\% reduction in model size.

\begin{figure}[!t] 
    \centering
    \includegraphics[width=1\linewidth]{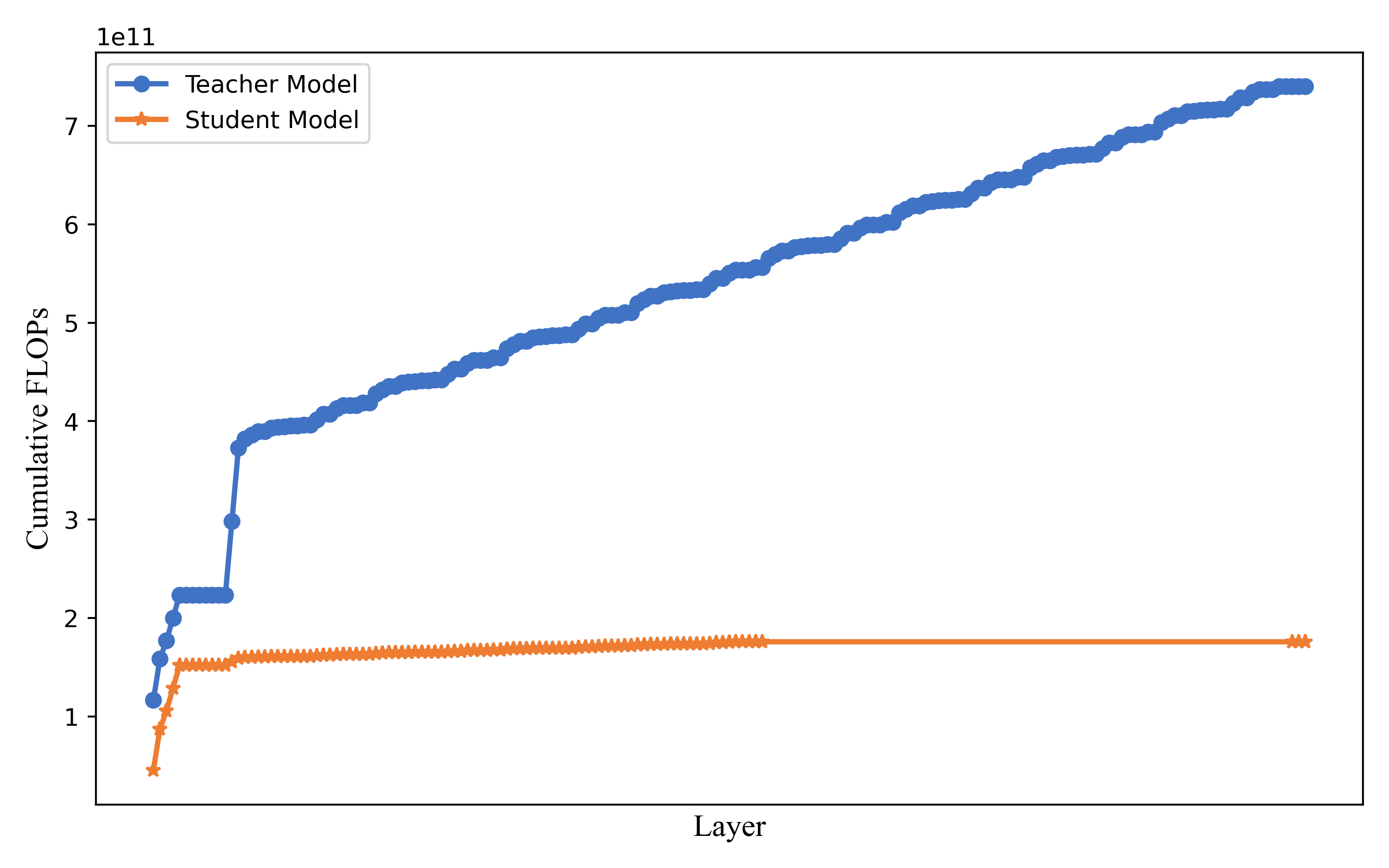}
    \caption{The cumulative FLOPs comparison between teacher and student models.}
    \label{fig:flops}
\end{figure}
Despite the reduced parameter count, the student model maintains performance close to the teacher, demonstrating a strong balance between efficiency and accuracy. Knowledge distillation enables significant model compression with minimal performance degradation, allowing the student to deliver comparable results with lower computational demands.

Table~\ref{TUSZ results} and Table~\ref{AUBMC results} further support this claim by showing consistent class-wise performance across the TUSZ and AUBMC datasets. The student model closely matches the teacher in per-class accuracy, confirming the effectiveness of the distillation framework in preserving fine-grained classification ability.

To assess computational efficiency, Fig.~\ref{fig:flops} compares cumulative FLOPs across model layers. The teacher model exceeds $7 \times 10^{11}$ FLOPs, while the student remains below $2 \times 10^{11}$, highlighting a substantial reduction in computational cost. This efficiency is especially notable in deeper layers, where the student model significantly reduces FLOPs without compromising performance.

\begin{figure*}[!t] 
    \centering
    \includegraphics[width=1\linewidth]{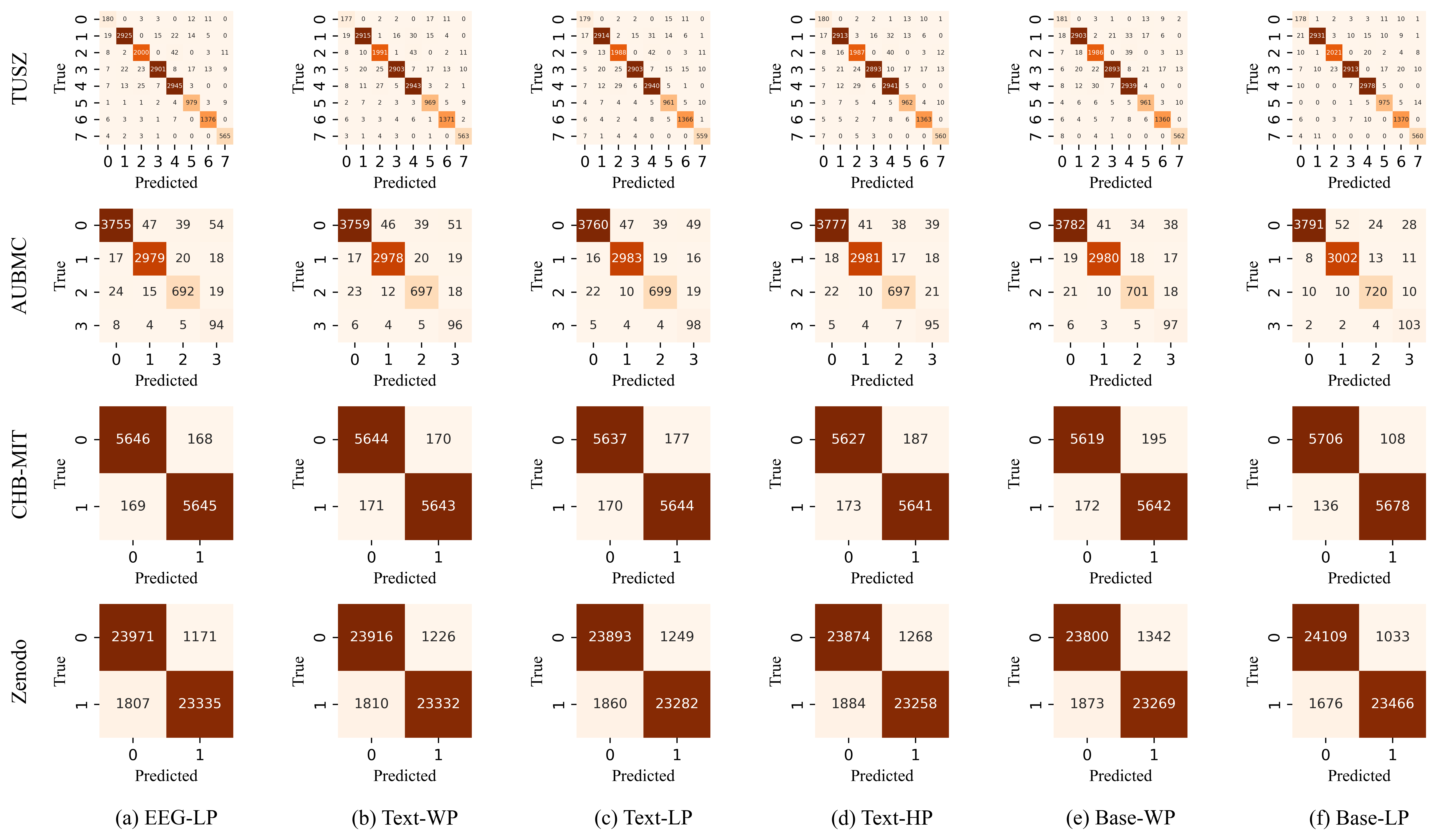}
    \caption{Confusion matrix comparison of the proposed model and its ablated variants across four EEG datasets}
    \label{fig:cm}
\end{figure*}
\begin{figure*}[t] 
    \centering
    \includegraphics[width=1\linewidth]{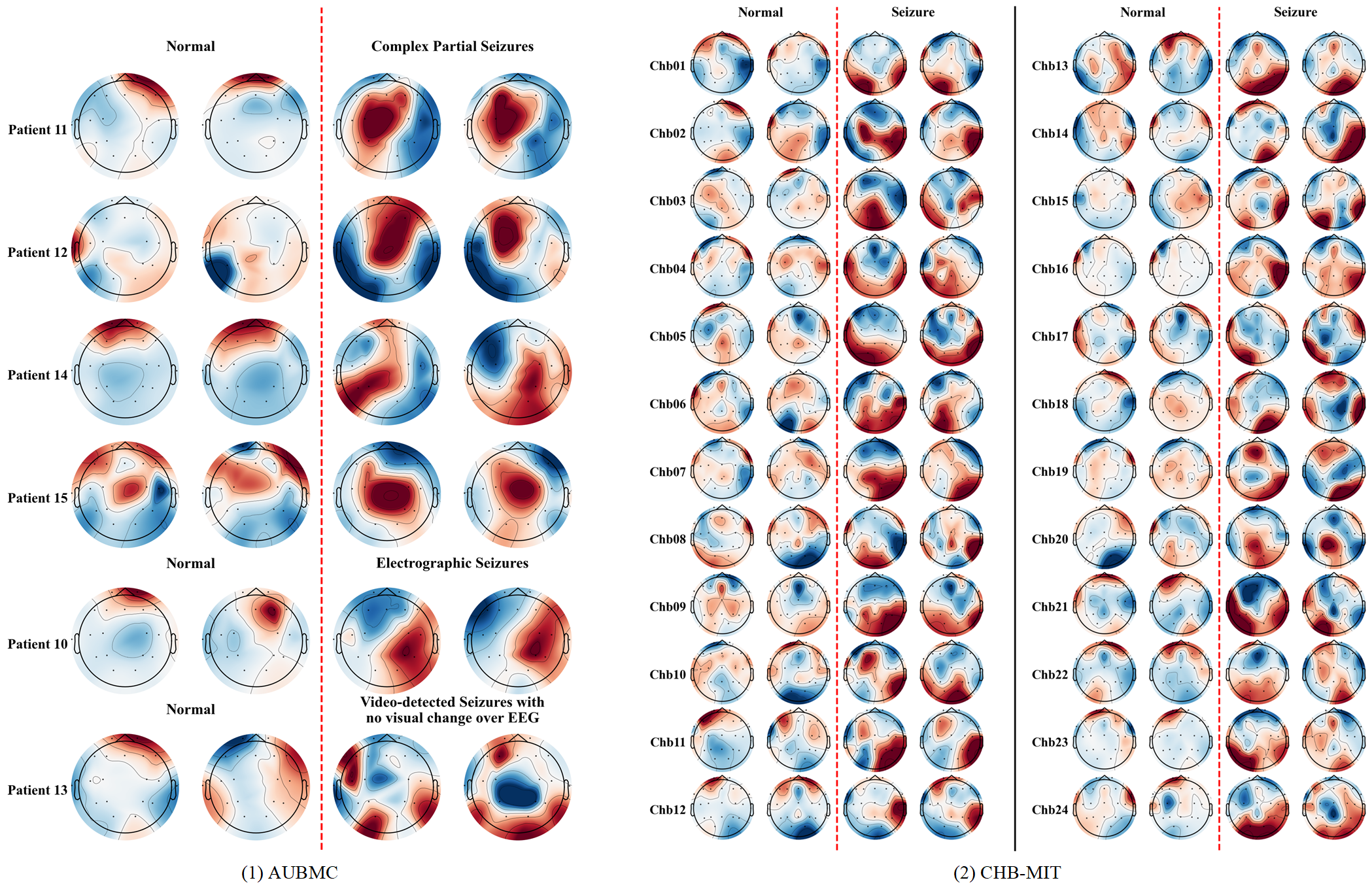}
    \caption{The EEG Channel Activation Maps (ECAMs) of all subjects obtained from the DistilCLIP-EEG model.}
    \label{fig:ECAMs}
\end{figure*}

Overall, these results demonstrate that the proposed knowledge distillation strategy enables the student model to generalize effectively while dramatically reducing model size, FLOPs, and inference cost. This efficiency makes it highly suitable for real-time and edge-based seizure detection in performance-constrained environments.

The classification performance of DistilCLIP-EEG was further evaluated using Area Under the Curve (AUC) metrics, including macro- and micro-averaged scores for multi-class datasets (TUSZ and AUBMC) and standard AUC for binary-class datasets (CHB-MIT and Zenodo). As shown in Table~\ref{tab:AUC}, both teacher and student models consistently achieved high AUC values, validating the effectiveness of the proposed framework.

A comprehensive ablation study was conducted on the teacher model across all four datasets to assess the impact of key components, particularly the learnable prompts integrated into both EEG and text encoders. Table~\ref{tab:ablation} shows that the full model (Base-LP) outperforms all ablated variants. Removing prompts from either encoder leads to noticeable performance degradation, while eliminating prompt learning entirely (Base-WP) results in the lowest accuracy—most significantly on the Zenodo dataset—highlighting the importance of prompt-based adaptation, especially under low-resource conditions.

Confusion matrices in Fig.~\ref{fig:cm} further illustrate the class-wise performance of the full and ablated models. The full model exhibits more balanced and accurate classification with fewer misclassifications, whereas ablated versions show increased confusion between similar classes, particularly in multi-class settings.

To enhance interpretability, we visualized EEG Channel Activation Maps (ECAMs) from the trained model, as shown in Fig.~\ref{fig:ECAMs}. ECAMs highlight spatial activation patterns across EEG channels for both normal and seizure events in the AUBMC and CHB-MIT datasets. Red regions denote higher activation, indicating greater model attention and potential seizure onset zones. These physiologically meaningful visualizations confirm that DistilCLIP-EEG not only performs well but also offers interpretable insights into its decision process, supporting its potential application in clinical settings.

\section{Conclusion} 
\label{sec:Conclusion}
This study introduces DistilCLIP-EEG, a multimodal epileptic seizure detection model that integrates EEG signals and textual descriptions using a Conformer-based EEG encoder, BERT-LP for text encoding, and knowledge distillation. The model achieves superior performance on benchmark datasets in terms of accuracy, F1 score, and AUC, effectively distinguishing seizure types and handling EEG complexity. Prompt learning enhances adaptability across datasets, while high AUC scores and confusion matrix results confirm its precision and robustness. The model also provides interpretable insights, supporting its potential clinical utility.

However, limitations remain. The use of template-based text may reduce adaptability to diverse clinical annotations. Additionally, the evaluation relies on cross-validation without subject-independent testing, limiting insight into generalization across unseen patients.

Future work should incorporate diverse demographic data through international collaboration and pursue subject-independent validation. Optimizing for real-time deployment via lightweight architectures and hardware acceleration is critical for clinical use. Integrating additional modalities (e.g., fMRI, genetic data) and developing personalized models based on patient-specific EEG and medical history may further improve performance. Clinical trials and integration into epilepsy monitoring systems are vital steps toward practical implementation.



\section*{References}
\bibliographystyle{IEEEtran}
\bibliography{reference}

\end{document}